\documentclass[10pt,twocolumn,letterpaper]{article}

\usepackage[pagenumbers]{iccv} % To force page numbers, e.g. for an arXiv version

\usepackage{epsfig}
\usepackage{graphicx}
\usepackage{amsmath}
\usepackage{amssymb}
\usepackage[acronym]{glossaries}

\usepackage{booktabs}      % for \toprule, \midrule, \bottomrule
\usepackage{multirow}      % for \multirow
\usepackage{array}         % for advanced column formatting (optional but recommended)
\usepackage{xcolor}        % for row coloring with \rowcolor
\usepackage{colortbl}      % for \rowcolor to work properly in tables
\usepackage{pifont}        % for checkmark (\cmark) and crossmark (\xmark)
\usepackage{makecell}      % for \shortstack and better line breaks in cells
\usepackage{caption}       % for customizing captions
\usepackage{float}         % for positioning (optional, if using [t], [h], etc.)
\usepackage[accsupp]{axessibility}  % Improves PDF readability for those with disabilities.

\newacronym{bce}{BCE}{binary cross-entropy}
\newacronym{ood}{OoD}{out-of-distribution}
\newacronym{oe}{OE}{outlier exposure}
\newacronym{edl}{EDL}{evidential deep learning}
\newacronym{id}{ID}{in-distribution}
\newacronym{nn}{NN}{neural network}
\newacronym{aspp}{ASPP}{atrous spatial pyramid pooling}
\newacronym{cnn}{CNN}{convolutional neural network}
\newacronym{sl}{SL}{subjective logic}
\newacronym{ap}{AP}{average precision}
\newacronym{fpr}{FPR}{false positive rate}

\definecolor{iccvblue}{rgb}{0.21,0.49,0.74}
\usepackage[pagebackref,breaklinks,colorlinks,allcolors=iccvblue]{hyperref}
\def\paperID{3} % *** Enter the Paper ID here
\def\confName{ICCV}
\def\confYear{2025}

\title{Uncertainty-Aware Likelihood Ratio Estimation for Pixel-Wise Out-of-Distribution Detection}

\author{
Marc Hölle \quad Walter Kellermann \quad Vasileios Belagiannis \\
Friedrich-Alexander-Universität Erlangen-Nürnberg, Germany\\
{\tt\small \{marc.hoelle, walter.kellermann, vasileios.belagiannis\}@fau.de}
}

\begin{document}

\twocolumn[{%
\renewcommand\twocolumn[1][]{#1}%
\maketitle
\begin{center}
    \captionsetup{type=figure}
    \vspace{-0.2in}
    \begin{minipage}[b]{0.01662\textwidth}
        \includegraphics[width=\linewidth]{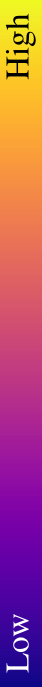} % Adjust height to span both rows
    \end{minipage}%
    \hspace{0.03em}
    \begin{minipage}[b]{0.97\textwidth}
        \begin{minipage}[b]{0.245\textwidth}
            \includegraphics[width=\linewidth]{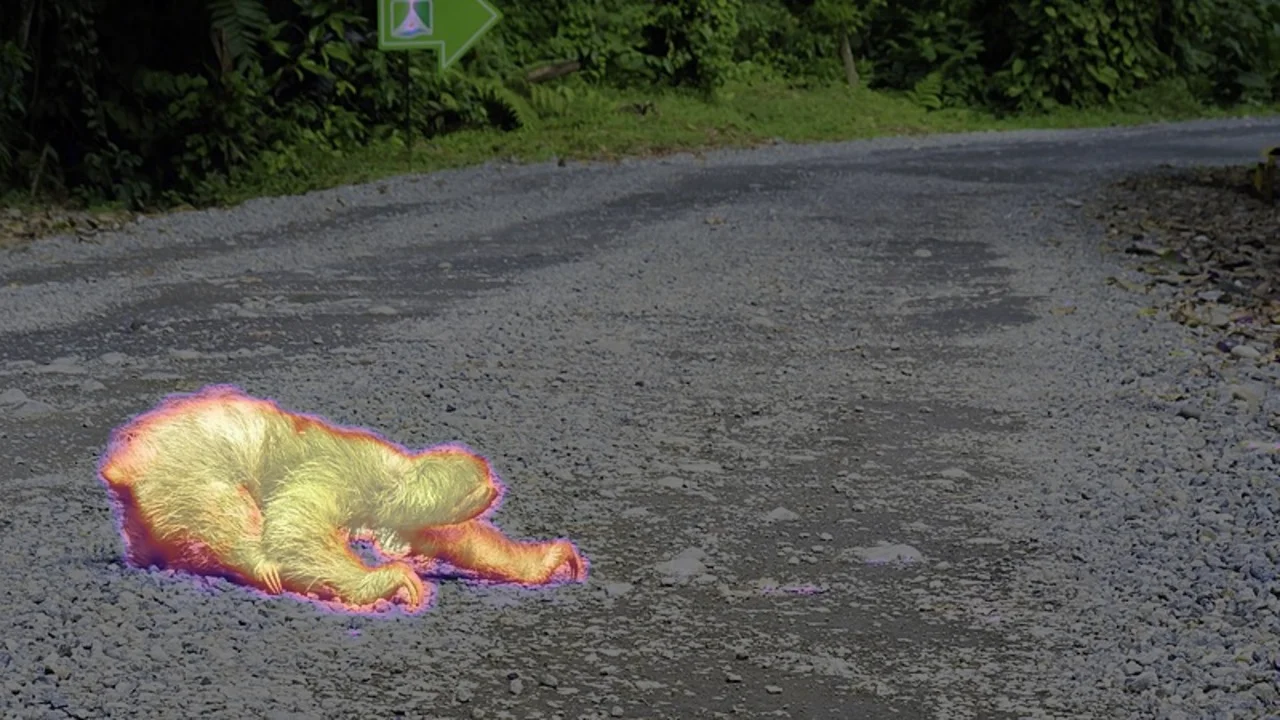}
        \end{minipage}
        \begin{minipage}[b]{0.245\textwidth}
            \includegraphics[width=\linewidth]{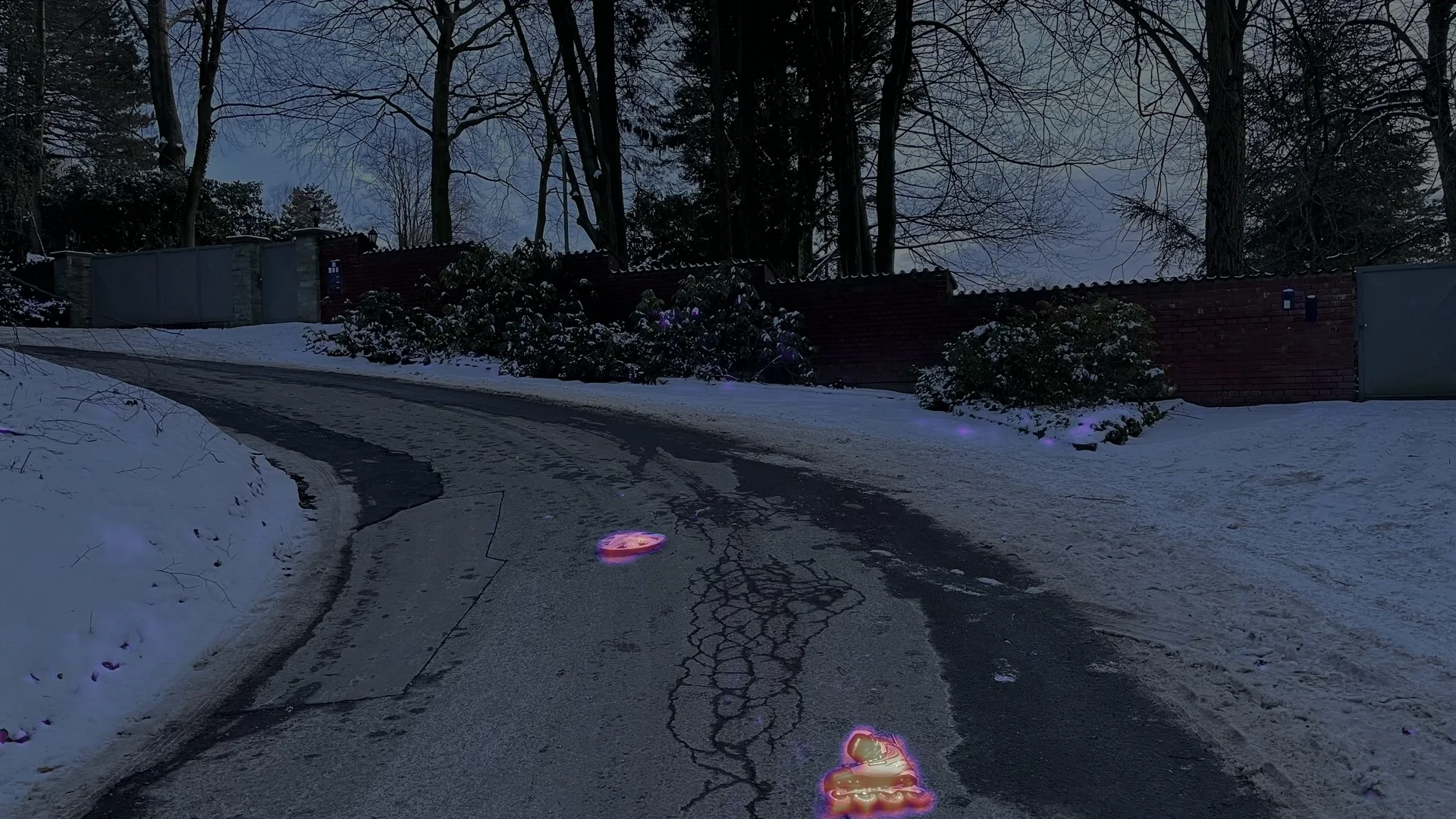}
        \end{minipage}
        \begin{minipage}[b]{0.245\textwidth}
            \includegraphics[width=\linewidth]{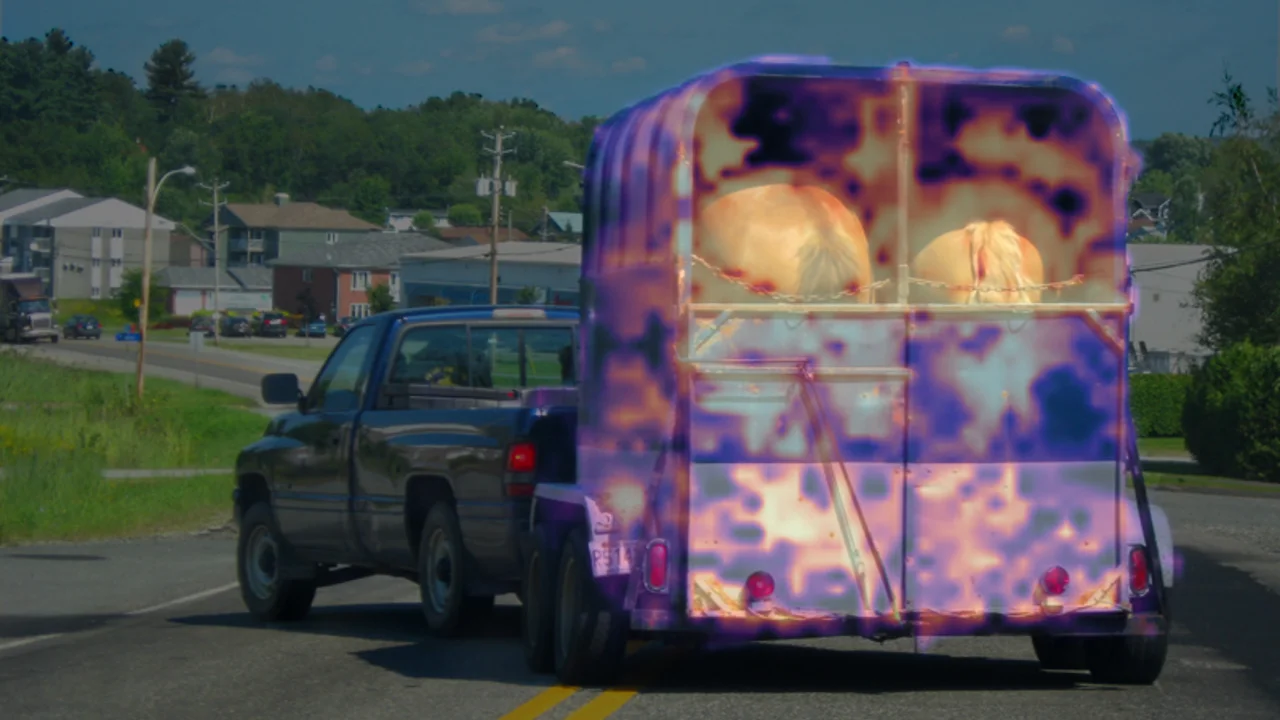}
        \end{minipage}
        \begin{minipage}[b]{0.245\textwidth}
            \includegraphics[width=\linewidth]{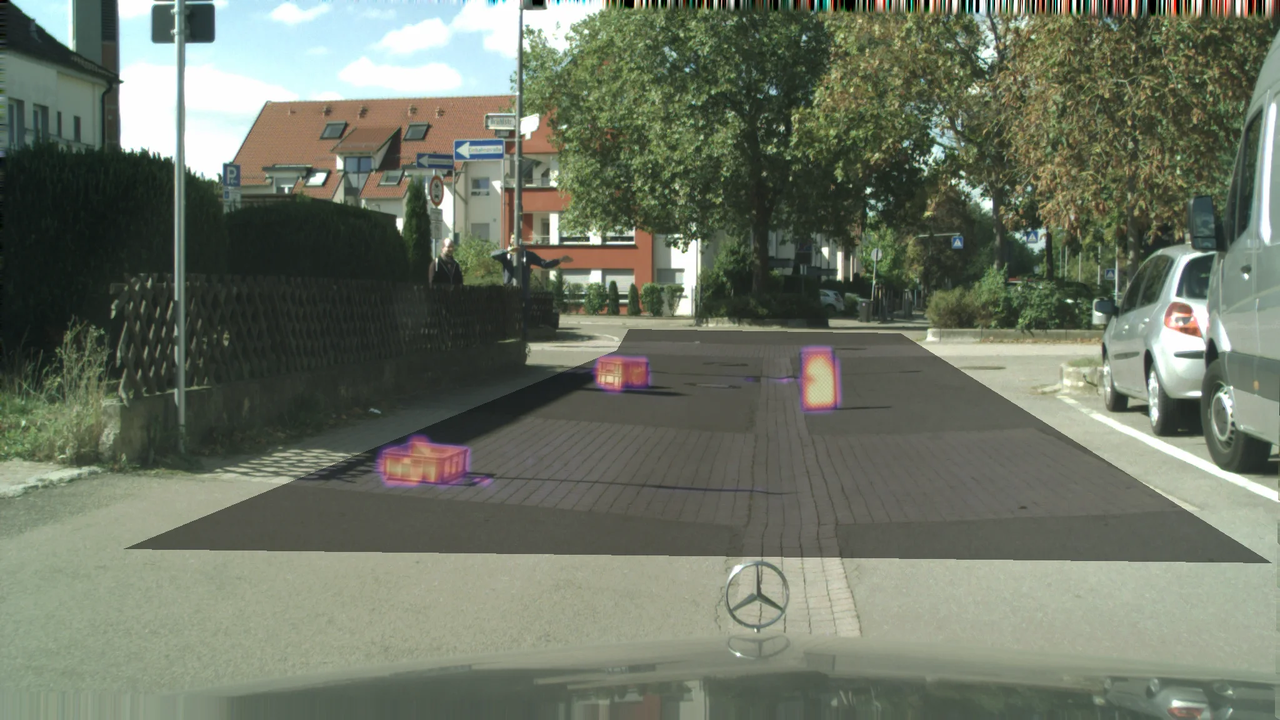}
        \end{minipage}
        
        \vspace{0.1em}
        
        \begin{minipage}[b]{0.245\textwidth}
            \includegraphics[width=\linewidth]{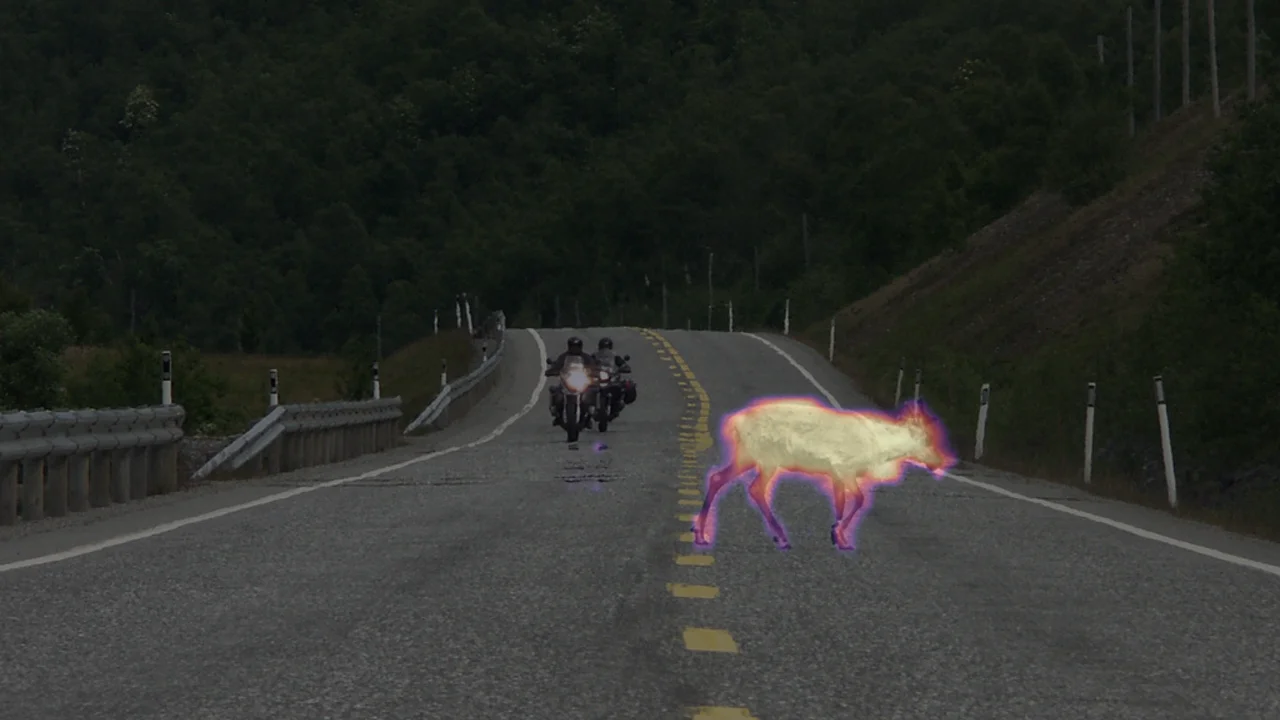}
        \end{minipage}
        \begin{minipage}[b]{0.245\textwidth}
            \includegraphics[width=\linewidth]{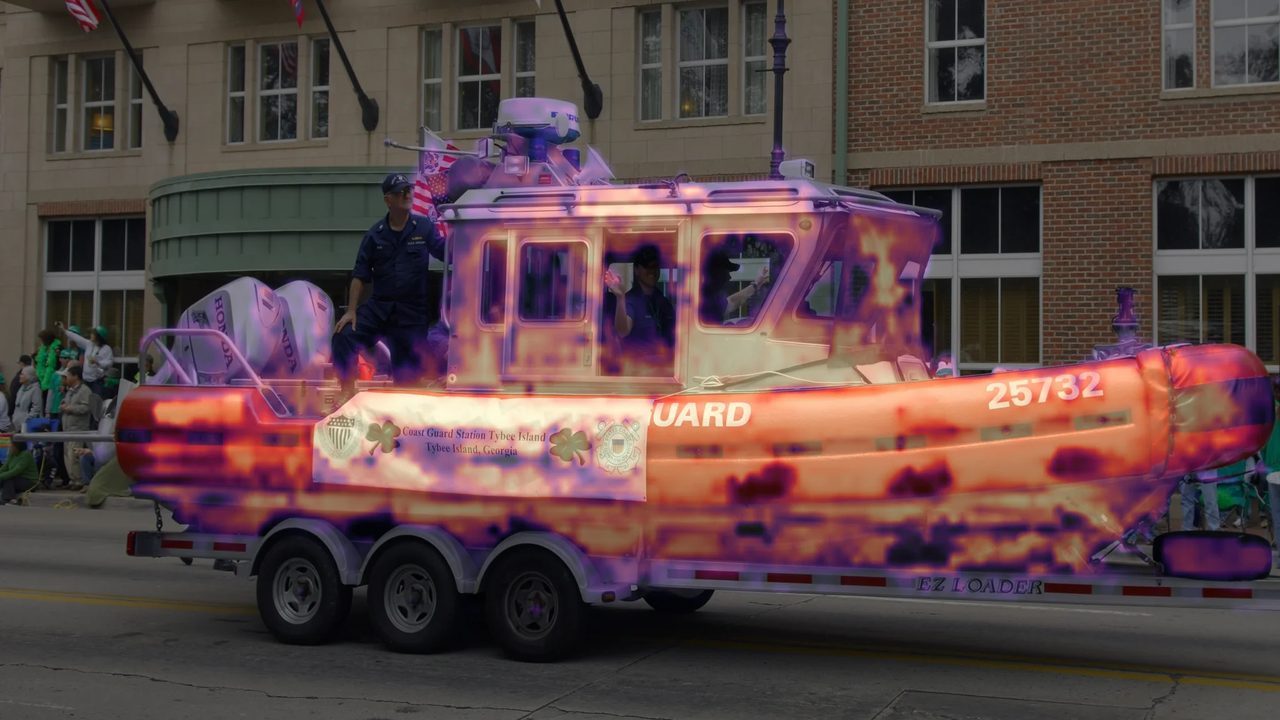}
        \end{minipage}
        \begin{minipage}[b]{0.245\textwidth}
            \includegraphics[width=\linewidth]{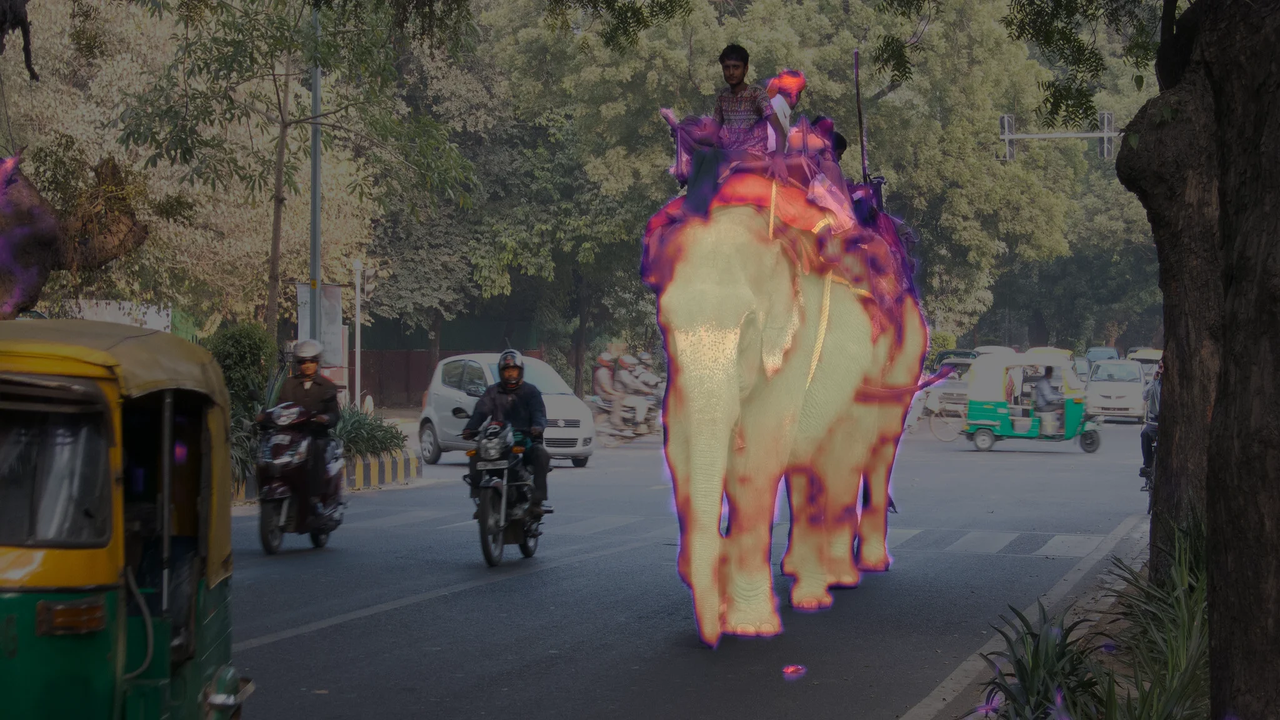}
        \end{minipage}
        \begin{minipage}[b]{0.245\textwidth}
            \includegraphics[width=\linewidth]{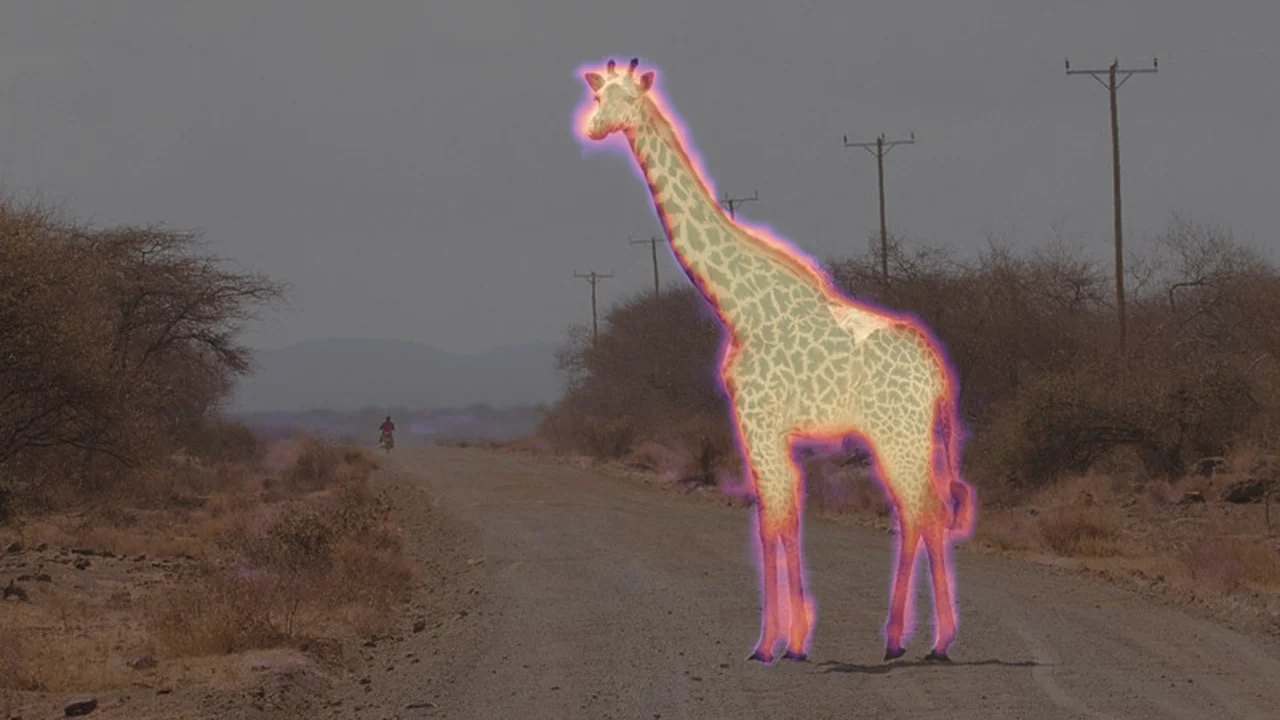}
        \end{minipage}
    \end{minipage}
    
    \captionof{figure}{\textbf{Visualisations for pixel-wise likelihood ratio.} Overlay of predicted pixel-wise likelihood ratios on images with unknown objects from diverse driving scenes. Our method successfully detects unknown objects without false detections on in-distribution regions. Images are taken from the Segment-Me-If-You-Can~\cite{chan2021segmentmeifyoucan} and Lost and Found no Known~\cite{pinggera2016lost} datasets.}
    \label{fig:teaser}
\end{center}
}]
\maketitle
\begin{abstract}
   Semantic segmentation models trained on known object classes often fail in real-world autonomous driving scenarios by confidently misclassifying unknown objects. While pixel-wise out-of-distribution detection can identify unknown objects, existing methods struggle in complex scenes where rare object classes are often confused with truly unknown objects. We introduce an uncertainty-aware likelihood ratio estimation method that addresses these limitations. Our approach uses an evidential classifier within a likelihood ratio test to distinguish between known and unknown pixel features from a semantic segmentation model, while explicitly accounting for uncertainty. Instead of producing point estimates, our method outputs probability distributions that capture uncertainty from both rare training examples and imperfect synthetic outliers. We show that by incorporating uncertainty in this way, outlier exposure can be leveraged more effectively. Evaluated on five standard benchmark datasets, our method achieves the lowest average false positive rate (2.5\%) among state-of-the-art while maintaining high average precision (90.91\%) and incurring only negligible computational overhead.
   Code is available at \url{https://github.com/glasbruch/ULRE}.
\end{abstract}

%------------------------------------------------------------------------
\section{Introduction}
Semantic segmentation models are typically trained under the closed-set assumption, where only a fixed set of known semantic classes is considered. While these models perform well on curated datasets, they often fail in real-world open-set scenarios~\cite{bendale2016towards, scheirer2012toward}, confidently misclassifying unknown objects as one of the known classes~\cite{hein2019relu}. Pixel-wise \gls{ood} detection~\cite{tian2022pixel, liu2023residual} addresses this limitation by identifying unknown objects at the pixel level, thereby complementing the standard \gls{id} segmentation with an additional \gls{ood} segmentation map. However, pixel-wise \gls{ood} detection remains particularly challenging in complex urban driving scenes. For example, a pixel labelled as "car" may correspond to distinct object parts such as a windshield, door or tyre, leading to high intra-class variability. Additionally, training data typically exhibits long-tailed class distributions, where common classes like "road" or "sky" are overrepresented, while others such as "bicycle" appear infrequently. These factors introduce uncertainty into the segmentation task, making it difficult to reliably distinguish between truly unknown objects and rare \gls{id} classes. %This imbalance further complicates the task of reliable \gls{ood} detection across all semantic regions.

Recent methods~\cite{liu2023residual, nayal2023rba, nayal2024likelihood, delic2024outlier} have addressed these challenges using \gls{oe}~\cite{hendrycks2018deep}, which augments the training of \gls{id} segmentation models with synthetic proxy \gls{ood} objects and encourages the model to produce highly uncertain predictions across the known classes for these objects. However, these approaches~\cite{liu2023residual, nayal2023rba} struggle to disentangle different sources of uncertainty. In particular, they fail to distinguish between uncertainty due to intrinsic class ambiguity or rare tail classes~\cite{nayal2023rba}, which occur infrequently in the training data, and uncertainty induced by true \gls{ood} objects. As a result, such methods often misclassify tail classes as \gls{ood}, leading to false detections. To overcome these limitations, a different approach is necessary that does not rely on uncertainty estimates or overconfident predictions from the \gls{id} segmentation model.

In this work, we introduce \textit{uncertainty-aware likelihood ratio estimation} for pixel-wise \gls{ood} detection. Our approach extracts dense intermediate feature representations from a pre-trained \gls{id} semantic segmentation model and performs a likelihood ratio test to determine whether a given pixel belongs to an \gls{ood} object. However, rather than explicitly estimating the feature distributions of \gls{id} and \gls{ood} pixels~\cite{galesso2024diffusion}, we propose a likelihood ratio estimation approach by training a binary \gls{nn} classifier to directly distinguish between \gls{id} and synthetic proxy \gls{ood} features, obtained by using \gls{oe}. Crucially, unlike previous work~\cite{nayal2024likelihood}, we explicitly acknowledge that these proxy features are only approximations of true \gls{ood} features and may introduce distributional biases. To address these biases, we propose an uncertainty-aware estimation scheme based on \gls{edl}. Therefore, instead of producing point estimates, our method outputs a distribution over likelihood ratio estimates, capturing epistemic uncertainty arising from long-tailed \gls{id} distributions and synthetic proxy \gls{ood} samples. This makes our approach more robust, particularly when synthetic features differ from real-world \gls{ood} instances. Importantly, our method maintains \gls{id} segmentation performance while adding only negligible computational overhead.

We first evaluate our approach on an illustrative estimation problem, showing that our uncertainty-aware estimator is less susceptible to overconfident extrapolation beyond the support of the training data compared to a standard estimator. We then assess performance on five established benchmarks, including Fishyscapes~\cite{blum2021fishyscapes}, Road Anomaly~\cite{lis2019detecting}, and Segment-Me-If-You-Can (SMIYC)~\cite{chan2021segmentmeifyoucan}. Across these datasets, our method achieves competitive results with state-of-the-art approaches, while consistently outperforming a standard likelihood ratio estimator. Our contributions can be summarized as follows:
\begin{itemize}
    \item We introduce uncertainty-aware likelihood ratio estimation that mitigates biases in synthetic \gls{ood} features; 
    \item We find that accounting for uncertainty enables more effective use of synthetic outliers, providing a new perspective for pixel-wise \gls{ood} detection with \gls{oe}; 
    \item Averaged across all five standard datasets our approach achieves the lowest false positive rate of 2.5~\% among state-of-the-art, while obtaining a strong average precision of 90.91~\%, demonstrating superior stability across diverse driving scene contexts and unknown objects.
\end{itemize}

%------------------------------------------------------------------------
\section{Related Work}
We review prior work on pixel-wise \gls{ood} detection (Sec. \ref{sec:related:ood}), likelihood ratio estimation with \glspl{nn} (Sec. \ref{sec:related:lr}) and \gls{edl} for \gls{ood} detection (Sec. \ref{sec:related:edl}) and relate it to specific features of the proposed scheme.

\subsection{Pixel-Wise Out-of-Distribution Detection}
\label{sec:related:ood}
Pixel-wise \gls{ood} detection methods can be categorized along two key axes: (1) whether the \gls{id} semantic segmentation model is kept frozen or retrained, and (2) whether \gls{oe} is used during training.

\paragraph{Freeze or Retrain} 
Pixel-wise \gls{ood} detection methods typically either freeze the underlying model or feature extractor~\cite{nayal2024likelihood, liu2023residual, di2021pixel, grcic2024dense, vojivr2024pixood, hornauer2023out}, or retrain parts of, or even the entire, \gls{id} model~\cite{tian2022pixel, chan2021entropy, grcic2022densehybrid, nayal2023rba}. However, retraining can degrade the performance of the original \gls{id} segmentation task~\cite{liu2023residual}. To mitigate this issue, we perform \gls{ood} detection without altering the parameters of the semantic segmentation model. Instead, we extract intermediate dense feature maps from the backbone of a pre-trained segmentation network. As a result, our method remains agnostic to the underlying model architecture and training protocol.

\paragraph{Without Outlier Exposure}
Pixel-wise \gls{ood} detection approaches that do not use \gls{oe} typically rely on uncertainty measures derived from the \gls{id} predictive distribution~\cite{vojivr2024pixood}, such as entropy~\cite{chan2021entropy}, or distance-based metrics in feature space~\cite{lee2018simple, galesso2023far, rabinowitz2025ghost}. Generative models have also been employed to directly model the distribution of \gls{id} features and detect \gls{ood} pixels by identifying low-density regions in the feature space~\cite{galesso2024diffusion}. However, relying solely on the estimated density of \gls{id} data has been shown to be unreliable for \gls{ood} detection~\cite{zhangfalsehoods}, particularly when \gls{id} and \gls{ood} features are similar, leading to misestimation~\cite{zhang2021understanding}. To address these shortcomings, we extract intermediate features from a frozen \gls{id} segmentation model and estimate the likelihood ratio between the distributions of \gls{id} and proxy \gls{ood} features. This enables more reliable pixel-wise \gls{ood} detection, as it shifts the focus from accurately modelling each distribution to directly distinguishing between them, making the approach more robust to density estimation errors in high-dimensional spaces~\cite{sugiyama2012density, ren2019likelihood}.

\paragraph{With Outlier Exposure} 
\Gls{ood} detection methods employing \gls{oe} leverage auxiliary outlier data to improve detection performance~\cite{hendrycks2018deep}. Specifically, for pixel-wise \gls{ood} detection, synthetic \gls{ood} objects are inserted into \gls{id} training images~\cite{tian2022pixel}. Furthermore, recent mask-based segmentation approaches~\cite{cheng2021per} have been adapted for \gls{ood} detection to move from pixel-wise to region-level classification \cite{nayal2023rba, grcic2023advantages, delic2024outlier, rai2023unmasking}, training models to predict object masks first and then classify each mask instead of individual pixels. While these methods achieve impressive segmentation performance on medium to large unknown objects, they may incorrectly label entire regions as \gls{ood} and miss small \gls{ood} objects~\cite{nayal2024likelihood, nayal2023rba}. To reduce false detections, we perform pixel-wise detection and explicitly address the mismatch between true \gls{ood} features and proxy \gls{ood} features by modelling uncertainty.

\subsection{Likelihood Ratio Estimation}
\label{sec:related:lr}
The likelihood ratio plays a central role in statistical inference and hypothesis testing, but its direct computation is often intractable when the underlying probability densities are unknown or computationally expensive to evaluate~\cite{sugiyama2012density}. Recent work has shown that data-driven approaches using \gls{nn}-based classifiers can approximate the likelihood ratio effectively~\cite{rizvi2024learning}. In the context of image-wise \gls{ood} detection, likelihood ratio estimation has been proposed as a principled detection strategy~\cite{zhangfalsehoods}. Several existing methods can be interpreted as performing \gls{ood} detection using a proxy \gls{ood} distribution~\cite{bitterwolf2022breaking}. However, only a few works have extended this formulation to the pixel-wise setting~\cite{nayal2024likelihood, vojivr2024pixood}. Moreover, none of these approaches explicitly account for the intrinsic uncertainty that arises when estimating the likelihood ratio using samples from a proxy \gls{ood} distribution.

\subsection{Evidential Deep Learning}
\label{sec:related:edl}
\Gls{edl} \cite{sensoy2018evidential, ulmer2023prior} extends traditional deep learning by modelling uncertainty through distributions over predictions, allowing the network to quantify uncertainty without requiring multiple forward passes \cite{gal2016dropout, de2025diffusion}, ensembles \cite{lakshminarayanan2017simple}, additional modules~\cite{lahlou2023deup, hornauer2023heatmap} or back-propagation during inference~\cite{hornauer2025revisiting}. Existing methods leveraging \gls{edl} for \gls{ood} detection utilise the estimated uncertainty over \gls{id} classes to detect \gls{ood} instances as samples or regions with high epistemic uncertainty \cite{hammam2023identifying, ancha2024deep}. In contrast, we propose to use the uncertainty-aware training of \gls{edl} to train a binary classifier with \gls{oe} to reduce overconfident extrapolation beyond the training set. %and potential over-fitting.

%------------------------------------------------------------------------
\section{Methodology}
\label{sec:method}
\glsresetall

We present a pixel-wise \gls{ood} detection approach that takes an image $\mathbf{I} \in \mathcal{I} \subset \mathbb{R}^{H \times W \times 3}$ to predict the \gls{ood} map $\hat{\mathbf{y}}^{\text{out}} \in \mathcal{Y}^{\text{out}} \subset (0,1)^ {H \times W}$, identifying pixels belonging to objects whose semantic class is not among the $C$ classes of the \gls{id} training dataset. To this end, we assume a pre-trained \gls{id} semantic segmentation model that classifies each pixel into a predefined closed-set of $C$ classes by predicting the segmentation map $\hat{\mathbf{y}}^{\text{in}} \in \mathcal{Y}^{\text{in}} \subset (0,1)^ {H \times W \times C}$. To identify \gls{ood} pixels, we extract dense intermediate feature maps $\mathbf{x} \in \mathcal{X} \subset \mathbb{R}^{H \times W \times D}$ from the pre-trained model. Specifically, for each pixel $i \in [H\cdot W]$ we use the corresponding feature vector $\mathbf{x}_i \in \mathbb{R}^{D}$ to determine whether it belongs to an \gls{ood} object.  

We propose a pixel-wise \gls{ood} approach by formulating the task as a likelihood ratio test (Sec. \ref{sec:method:llr}), where we approximate the ratio between the true \gls{id} feature distribution  $p_{\text{in}}(\mathbf{x}_i)$ and a proxy \gls{ood} $\tilde{p}_{\text{out}}(\mathbf{x}_i)$ (Sec. \ref{sec:method:proxy}). However, instead of estimating these individual densities, we introduce a binary \gls{nn} classifier $p_{\theta}(y_i^{\text{out}}|\mathbf{x}_i)$ to distinguish between \gls{id} and synthetic proxy \gls{ood} features (Sec. \ref{sec:method:estimator}). Since the choice of proxy \gls{ood} features $\mathbf{x}_i \sim \tilde{p}_{\text{out}}(\mathbf{x}_i)$ may introduce biases, we model the intrinsic uncertainty in this likelihood ratio estimation using \gls{edl} (Sec. \ref{sec:method:edl}). Finally, we propose an uncertainty-aware pixel-wise training scheme (Sec. \ref{sec:method:training}).

%Out-of-Distribution Detection and likelihood ratio
\subsection{Out-of-Distribution Detection as Likelihood Ratio Test}
\label{sec:method:llr}
We formulate pixel-wise \gls{ood} detection as a hypothesis test, where the null hypothesis $H_0$ assumes that feature vector $\mathbf{x}_i$ is drawn from an \gls{id} $p_{\text{in}}(\mathbf{x}_i)$, while the alternative hypothesis $H_1$ assumes that $\mathbf{x}_i$ is drawn from an \gls{ood} $p_{\text{out}}(\mathbf{x}_i)$ \cite{zhangfalsehoods}. 
According to the Neyman-Pearson lemma \cite{neyman1933ix}, a likelihood ratio test is the uniformly most powerful test for deciding whether $\mathbf{x}_i$ supports $H_0$ or $H_1$. In this test,  $\mathbf{x}_i$ is detected as \gls{ood} if the ratio
\begin{equation}
    \text{LR}(\mathbf{x}_i) = \frac{p_{\text{out}}(\mathbf{x}_i)}{p_{\text{in}}(\mathbf{x}_i)}
\end{equation}
exceeds a threshold $\text{LR}(\mathbf{x}_i) > \tau$.

\paragraph{Likelihood Ratio Trick} Notably, we can directly approximate $\text{LR}(\mathbf{x}_i)$ as
\begin{equation}
\label{eq:lr_trick}
    \begin{split}
    \frac{p_{\text{out}}(\mathbf{x}_i)}{p_{\text{in}}(\mathbf{x}_i)} 
    &= \frac{p(\mathbf{x}_i \mid O=1)}{p(\mathbf{x}_i \mid O=0)} \\
    &= \frac{P(O=1 \mid \mathbf{x}_i) p(\mathbf{x}_i)}{P(O=1)} 
    \Big/ \frac{P(O=0 \mid \mathbf{x}_i) p(\mathbf{x}_i)}{P(O=0)} \\
    &= \frac{P(O=1 \mid \mathbf{x}_i)}{P(O=0 \mid \mathbf{x}_i)},
\end{split}
\end{equation}
where $O$ is a Bernoulli random variable determining whether $\mathbf{x}_i$ is drawn from $p_{\text{in}}(\mathbf{x}_i)$ or $p_{\text{out}}(\mathbf{x}_i)$. The feature distribution is described by 
\begin{equation}
\begin{split}
    p(\mathbf{x}_i) &= p(\mathbf{x}_i \mid O=1)P(O=1) \\
    &\quad + p(\mathbf{x}_i \mid O=0)P(O=0),
\end{split}
\end{equation}
with a uniform prior $P(O=0)=P(O=1)$.\footnote{A non-uniform prior $P(O=0)\neq P(O=1)$ introduces a constant scaling factor $P(O=0)/P(O=1)$, which can be absorbed into the decision threshold $\tau$.}
This formulation allows us to approximate the likelihood ratio without explicitly estimating the individual densities $p_{\text{in}}(\mathbf{x}_i)$ and $p_{\text{out}}(\mathbf{x}_i)$. 

In practice, however, we only have access to features $\mathbf{x}_i \sim p_{\text{in}}(\mathbf{x}_i)$ from the \gls{id} training dataset, while the true \gls{ood} $p_{\text{out}}(\mathbf{x}_i)$ is unknown as it has to represent features from all possible semantic classes outside the \gls{id} training dataset. To address this limitation, we use \gls{oe} \cite{hendrycks2018deep, zhang2023mixture} to obtain features from a proxy \gls{ood} $\tilde{p}_{\text{out}}(\mathbf{x}_i)$. %We propose to use \gls{oe} to obtain potentially biased samples from a proxy distribution $\tilde{p}_{\text{out}}$.

\subsection{Proxy Out-of-Distribution}
\label{sec:method:proxy}

\Gls{oe} improves the distinction between \gls{id} and \gls{ood} samples by incorporating auxiliary outlier data during training. Following prior work on pixel-wise \gls{ood} detection \cite{liu2023residual, nayal2023rba, tian2022pixel}, we utilise \gls{oe} by inserting synthetic \gls{ood} objects into \gls{id} training images. From these modified images, we extract feature maps $\mathbf{x}$ using a pre-trained segmentation model. The feature vectors $\mathbf{x}_i$ corresponding to the synthetic \gls{ood} object are then treated as samples from a proxy \gls{ood} $\tilde{p}_{\text{out}}(\mathbf{x}_i)$.

%To generate features from $\tilde{p}_{\text{out}}(\mathbf{x}_i)$, 
Specifically, we modify AnomalyMix \cite{tian2022pixel}, which cuts an \gls{ood} object from an outlier dataset, whose semantic class is not contained in the \gls{id} training dataset, and inserts it at a random location in an \gls{id} training image. Furthermore, we generate a pseudo \gls{ood} segmentation map $\mathbf{y}^{\text{out}} \in \mathcal{Y}^{\text{out}} \subset \{0,1\}^{H \times W}$ of the inserted \gls{ood} object. %Additionally, to enhance realism, we propose adjusting the inserted object's brightness and color to match the surrounding area in the \gls{id} image, blending it seamlessly into the context.

\subsection{Likelihood Ratio Estimator} 
\label{sec:method:estimator}
We propose to train a binary \gls{nn} classifier $p_{\theta}(y_i^{\text{out}}|\mathbf{x}_i)$, based on Eq. \ref{eq:lr_trick}, to distinguish between features from $p_{\text{in}}(\mathbf{x}_i)$ and $\tilde{p}_{\text{out}}(\mathbf{x}_i)$, assigning labels $y_i^{\text{out}}=0$ and $y_i^{\text{out}}=1$, respectively. Afterward, this classifier can be transformed into a likelihood ratio estimator as
\begin{equation}
\label{eq:lr_estimator}
    \hat{\text{LR}}(\mathbf{x}_i) = \frac{p_{\theta}(y_i^{\text{out}}=1|\mathbf{x}_i)}{p_{\theta}(y_i^{\text{out}}=0|\mathbf{x}_i)} = \frac{p_{\theta}(y_i^{\text{out}}=1|\mathbf{x}_i)}{1 - p_{\theta}(y_i^{\text{out}}=1|\mathbf{x}_i)}.
\end{equation}

%\paragraph{Overconfidence} 
We empirically observed that directly training a binary \gls{nn} classifier on \gls{id} and proxy \gls{ood} features using \gls{bce} leads to overconfident predictions on unseen data, particularly in low-density regions of feature space $\mathcal{X}$ where training data is scarce. This issue arises because standard \gls{bce} loss does not explicitly account for uncertainty in long-tailed \gls{id} and proxy \gls{ood} features, causing the model to assign overconfident probabilities to poorly represented regions. %Additionally, if the proxy \gls{ood} features do not sufficiently represent real-world \gls{ood} variability, the model may overfit to the specific characteristics of the training data, reducing its ability to generalize to true \gls{ood} features.

To mitigate this overconfidence, we propose next an uncertainty-aware estimation scheme leveraging \gls{edl} to model the intrinsic uncertainty in likelihood ratio estimation from proxy distribution features $\mathbf{x}_i \sim \tilde{p}_{\text{out}}(\mathbf{x}_i)$. %This approach mitigates overconfident extrapolation beyond the support of the training data. %and reduces potential overfitting.
%preventing overconfident extrapolation beyond the training data's support and mitigating potential overfitting.

\subsection{Uncertainty-Aware Likelihood Ratio Estimation}
\label{sec:method:edl}
We capture epistemic uncertainty arising from a long-tailed \gls{id} distribution and potentially biased proxy \gls{ood} features by incorporating \gls{edl} into likelihood ratio estimation. To this end, we first elaborate on the theoretical foundations of \gls{edl} in \gls{sl} and its connection to pixel-wise \gls{ood} detection. Finally, we describe how uncertainty can be quantified and how pixel-wise evidential classification is performed.

\paragraph{Subjective Logic} 
\Gls{edl} is based on \gls{sl} \cite{jsang2018subjective}, an extension of traditional probability theory that explicitly incorporates uncertainty. Instead of assigning single class probabilities $\mathbf{p}$, \gls{sl} models belief masses over categorical distributions using a Dirichlet distribution \cite{bishop2006pattern}, denoted as $\text{Dir}(\mathbf{p} \vert \boldsymbol{\alpha})$, where $\boldsymbol{\alpha}$ denotes the concentration parameters. For pixel-wise \gls{ood} detection, we therefore define a probability distribution $\text{Dir}(\mathbf{p}_i \vert \boldsymbol{\alpha}_i)$ over binary class probabilities $\mathbf{p}_i$ for each pixel $i$.

\paragraph{Pixel-Wise Evidential Model}
%For a classification task with $K$ classes\footnote{We distinguish $K$ from the number of \gls{id} classes $C$. In our case, $K=2$ for binary classification.}, given an input $\mathbf{x}_i$, the network provides evidence
For a binary classification task with $K=2$ classes, given the feature vector $\mathbf{x}_i$ extracted from the \gls{id} segmentation model, our network predicts the evidence \cite{sensoy2018evidential}
\begin{equation}
    \mathbf{e}_i = \exp(\mathbf{o}_i) \in \mathbb{R}^K_{+},
\end{equation}
where $\mathbf{o}_i$ represents the raw \gls{nn} classifier output (logits). Intuitively, $e_{i1}$ encodes the amount of evidence supporting the hypothesis that $\mathbf{x}_i$ is \gls{ood}. The belief mass assignment follows a Dirichlet distribution $\text{Dir}(\mathbf{p}_i \vert \boldsymbol{\alpha}_i)$ with parameters:
\begin{equation}
    \boldsymbol{\alpha}_i = \mathbf{e}_i + 1,
\end{equation}
ensuring $\boldsymbol{\alpha}_i \geq 1$ for a non-degenerate Dirichlet distribution. This formulation allows the model to express confidence and uncertainty in its predictions. When the model is confident, its predictions form a sharply peaked distribution. Conversely, when it is uncertain, the distribution becomes nearly uniform.
 
\paragraph{Measuring Uncertainty} The uncertainty can be quantified by the vacuity \cite{pandey2023learn}
\begin{equation}
    \nu_i = \frac{K}{S_i} = \frac{2}{S_i},
\end{equation}
with Dirichlet strength $S_i = \sum_{k=1}^K \alpha_{ik}$. Higher vacuity corresponds to greater uncertainty, indicating insufficient evidence for confident decision-making.

\paragraph{Pixel-Wise Evidential Classification}
We obtain binary class probabilities from the predicted Dirichlet distribution by calculating the expectation \cite{sensoy2018evidential}
\begin{equation}
\label{eq:expectation}
    \mathbb{E}_{\mathbf{p}_i \sim \text{Dir}(\mathbf{p}_i \vert \boldsymbol{\alpha}_i)} [\mathbf{p}_i] = \frac{\boldsymbol{\alpha}_i}{S_i}.
\end{equation}
From this, the binary classification probability follows as
\begin{equation}
    p_{\theta}(y_i^{\text{out}}=1|\mathbf{x}_i) = \frac{\alpha_{i1}}{S_i}.
\end{equation}

\subsection{Pixel-Wise Evidential Model Training}
\label{sec:method:training}
During training, we aim to maximise evidence for the correct class, minimise evidence for incorrect classes, and capture well-calibrated uncertainty. To achieve this, we employ the evidential log loss and evidence regularisation, which are combined with an annealing coefficient in the total loss.

\paragraph{Evidential Log Loss} 
For classification, we do not observe the true class probability $\mathbf{p}_i$ during training, only one-hot encoded class labels $\mathbf{y}^{\text{out}}_i \in \{0, 1\}^{K=2}$. Hence, we maximise the marginal likelihood (type II maximum likelihood)
\begin{equation}
    p(\mathbf{y}^{\text{out}}_i\vert \boldsymbol{\alpha}_i) = \int p(\mathbf{y}^{\text{out}}_i \vert \mathbf{p}_i) \text{Dir}(\mathbf{p}_i \vert \boldsymbol{\alpha}_i) d\mathbf{p}_i,
\end{equation}
and obtain the loss function \cite{pandey2023learn, sensoy2018evidential}
\begin{equation}
    \mathcal{L}^{\text{Log}}(\mathbf{x}_i, \mathbf{y}^{\text{out}}_i) = \sum_{k=1}^K y^{\text{out}}_k \left( \log(S_i) - \log(\alpha_{ik}) \right).
\end{equation}

\paragraph{Evidence Regularisation} To encourage the model to output low evidence for incorrect classes, we minimise the Kullback–Leibler divergence $\text{KL}(\cdot)$ between the Dirichlet distribution after removing correct evidence, denoted as $\text{Dir}(\mathbf{p}_i \vert \tilde{\boldsymbol{\alpha}}_i)$, and a uniform prior $\text{Dir}(\mathbf{p}_i \vert \mathbf{1})$ \cite{sensoy2018evidential}:
\begin{equation}
    \begin{split}
\mathcal{L}_{\text{reg}}^{\text{EDL}}(\mathbf{x}_i, \mathbf{y}^{\text{out}}_i) &= 
\text{KL}(\text{Dir}(\mathbf{p}_i \vert \tilde{\boldsymbol{\alpha}}_i) \parallel \text{Dir}(\mathbf{p}_i \vert \mathbf{1})) \\
&= \log \left( 
    \frac{\Gamma\left(\sum_{k=1}^K \tilde{\alpha}_{ik} \right)}
         {\Gamma(K) \prod_{k=1}^K \Gamma(\tilde{\alpha}_{ik})} 
\right) + {} \\
&\hspace*{-2em} \sum_{k=1}^K (\tilde{\alpha}_{ik} - 1) 
\left[ 
    \psi(\tilde{\alpha}_{ik}) 
    - \psi \left( \sum_{j=1}^K \tilde{\alpha}_{ij} \right) 
\right],
\end{split}
\end{equation}
where $\tilde{\boldsymbol{\alpha}}_i = \mathbf{y}^{\text{out}}_i + (\mathbf{1} - \mathbf{y}^{\text{out}}_i) \odot \boldsymbol{\alpha}_i = (\tilde{\alpha}_{i1}, \tilde{\alpha}_{i2})$ are the Dirichlet parameters after removing non-misleading evidence, $\mathbf{1}$ denotes a vector of $K$ ones, $\Gamma(\cdot)$ is the gamma function and $\psi(\cdot)$ is the digamma function. This regularisation prevents excessive confidence in incorrect predictions by encouraging the model to distribute evidence cautiously.

\paragraph{Total Loss} The overall loss combines the evidence log loss with the regularisation term
\begin{equation}
    \mathcal{L}(\mathbf{x}, \mathbf{y}^{\text{out}}) = \sum^N_{i=1} \mathcal{L}^{\text{Log}}(\mathbf{x}_i, \mathbf{y}_i^{\text{out}}) + \lambda_t \mathcal{L}_{\text{reg}}^{\text{EDL}}(\mathbf{x}_i, \mathbf{y}_i^{\text{out}}),
\end{equation} 
where $N=H\cdot W$, $t$ is the current training epoch and $\lambda_t = \min\left( 1, t/10\right) \in [0,1]$ is an annealing coefficient that gradually increases the strength of the KL divergence regularisation over the first 10 epochs. This prevents premature convergence to the uniform distribution for misclassified features \cite{sensoy2018evidential}.

%------------------------------------------------------------------------
\section{Experiments}

We evaluate our uncertainty-aware likelihood ratio estimator in two different settings. First, we present a proof-of-concept in a controlled experimental setup (Sec. \ref{sec:exp:univariate_gaussians}). Second, we evaluate our proposed method on standard pixel-wise \gls{ood} detection benchmarks for road anomaly and obstacle detection and compare it to the current state-of-the-art (Sec. \ref{sec:exp:OOD_detection}).
 
\subsection{Classification of Univariate Gaussians}
\label{sec:exp:univariate_gaussians}
\begin{figure}%[htbp]
  \centering
  \includegraphics[width=\columnwidth]{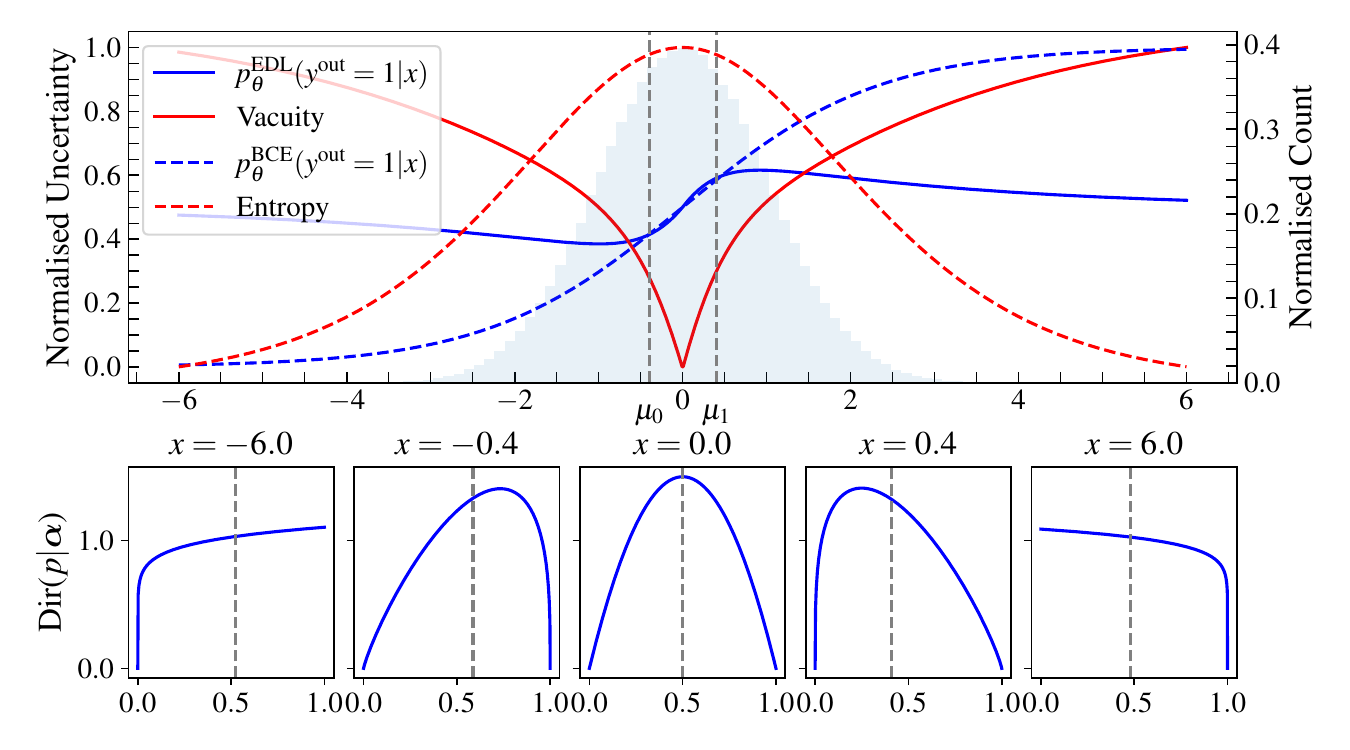} % Replace with your image filename
   \vspace{-2.5em}
  \caption{\textbf{Visualisation of univariate Gaussian classification.} \textbf{Top:} Comparison between an evidential classifier (EDL) and a standard binary classifier trained with binary cross-entropy (BCE), with corresponding uncertainty measures, vacuity and entropy, respectively. Shaded blue area indicates the training data density. \textbf{Bottom:} Predicted Dirichlet distributions $\text{Dir}(p|\boldsymbol{\alpha})$ at $x \in \{-6, -0.4, 0, 0.4, 6 \}$. The vertical dashed lines indicate the mean probability $p_{\theta}^{\text{EDL}}(y^{\text{out}}=1|x)$.}
  \label{fig:exp:univariate_gaussian}
\end{figure}

In order to investigate an analytically tractable classification problem, we consider two Gaussian distributions with unit variance and distinct means: $p_{\text{in}}(x)=\mathcal{N}(\mu_0=-0.4, 1)$ and $p_{\text{out}}(x)=\mathcal{N}(\mu_1=0.4, 1)$. This setup results in a binary classification task with significant class overlap and, consequently, high intrinsic uncertainty. 

\subsubsection{Experimental Setup} 
We train a fully connected \gls{nn} with a single hidden layer of size 16 and leaky ReLU activation. Two models are considered: a \textit{standard classifier} trained with \gls{bce} and sigmoid output activation, and our \textit{evidential classifier} trained using the loss described in Sec. \ref{sec:method:training} to distinguish between samples from $p_{\text{in}}(x)$ and $p_{\text{out}}(x)$. Both models are trained on $10^{6}$ samples drawn from $p_{\text{in}}(x)$ and $p_{\text{out}}(x)$ for 100 epochs, with early stopping based on validation performance.

\subsubsection{Analysis} Fig. \ref{fig:exp:univariate_gaussian} compares the predicted class probabilities of both classifiers. For reference, the class means, $\mu_0$ and $\mu_1$, are indicated by vertical dashed lines. Notably, the evidential classifier can express epistemic uncertainty through the shape of the predicted Dirichlet distribution, in contrast to the standard classifier's Bernoulli output, which conflates equal class likelihood with uncertainty. A uniform Bernoulli prediction of $p_{\theta}^{\text{BCE}}(y^{\text{out}}=1|x)=0.5$ indicates maximal uncertainty, without distinguishing whether this uncertainty originates from ambiguous data or insufficient evidence.%evidential model can express epistemic uncertainty through a uniform Dirichlet distribution. This stands in contrast to a uniform Bernoulli distribution, which only reflects equal class probabilities but cannot distinguish between a true lack of evidence and intrinsic class ambiguity.

Intuitively, in this overlapping-class scenario, a prediction close to $p_{\theta}(y^{\text{out}}=1|x) = 0.5$ can arise for two different reasons: (1) the input lies in a region of class overlap, where both classes are equally likely (aleatoric uncertainty), or (2) the model has insufficient evidence due to sparse training data (epistemic uncertainty). The evidential classifier distinguishes between these cases by capturing \textit{vacuity} i.e., lack of evidence. Specifically, it decreases evidence in low-density regions, while increasing it near $x=0$, where the training data is most concentrated. %Hereby, it increases uncertainty in low-density regions, while reducing it near $x=0$, where the training density is highest, while the class overlap is maximal and inherent uncertainty is unavoidable. 

This behaviour is expressed through the predicted Dirichlet distributions. At $x=0$, the predicted distribution has a mean of $p_{\theta}^{\text{EDL}}(y^{\text{out}}=1|x)=0.5$, reflecting maximal class ambiguity. However, the distribution is sharply peaked, indicating high confidence in this ambiguous but well-supported decision. As we move from the centre, the predicted distributions become flatter, reflecting increased epistemic uncertainty due to the lack of training data. This allows the evidential classifier to disentangle uncertainty caused by intrinsic ambiguity from that due to a lack of knowledge.

In contrast, uncertainty in a standard classifier is typically quantified using the binary entropy of its predicted probabilities. Its uncertainty peaks at $x=0$, but decreases in low-density regions, where it produces confident predictions, even for inputs far from the support of the training data, resulting in overconfident extrapolation. %Thus, a prediction of $p=0.5$ indicates maximal uncertainty without distinguishing whether this uncertainty originates from ambiguous data or insufficient evidence.

The evidential classifier's ability to distinguish between epistemic (model-based) and aleatoric (intrinsic) uncertainty allows it to avoid forcing confident predictions on ambiguous training samples. Instead, the model can reduce the strength of predicted evidence where appropriate, without compromising classification performance. We propose to leverage this property for likelihood ratio estimation using proxy distributions, where it is particularly beneficial, as both \gls{id} and \gls{ood} samples may be intrinsically ambiguous.

\subsection{Pixel-wise Out-of-Distribution Detection}
We evaluate our proposed method on five established benchmarks designed for real-world open-set autonomous driving scenarios, where reliable detection of unknown objects is critical.

\label{sec:exp:OOD_detection}
\subsubsection{Experimental Setup}
\paragraph{Datasets} We train our method using AnomalyMix~\cite{tian2022pixel}, which creates synthetic \gls{ood} examples by extracting objects from the COCO dataset~\cite{lin2014microsoft}, randomly resizing and pasting them at random locations within images from the \gls{id} training dataset, Cityscapes~\cite{cordts2016cityscapes}. We evaluate performance on the official SMIYC benchmark~\cite{chan2021segmentmeifyoucan}, which includes the Anomaly Track, Obstacle Track and Lost and Found no Known (LaF NoKnown) \cite{pinggera2016lost} datasets. Additionally, we include the Road Anomaly~\cite{lis2019detecting} and Fishyscapes Static~\cite{blum2021fishyscapes} datasets. Among these, Anomaly Track, Fishyscapes Static and Road Anomaly focus on semantically different objects like animals and unusual vehicles that are not present in the \gls{id} classes, while Obstacle Track and LaF NoKnown focus on small, previously unseen obstacles like boxes and debris on the road. %We train our method using AnomalyMix 

\paragraph{Evaluation Metrics} We evaluate the performance of our method using common pixel-wise segmentation metrics: \gls{ap} and \gls{fpr} at a true positive rate of 95\%.

\subsubsection{Implementation Details}

\paragraph{Feature Extractor} We use a frozen, self-supervised, pre-trained DINOv2 vision transformer with a patch size of 14 pixels (ViT-L/14 variant) ~\cite{oquab2023dinov2} and extract intermediate features from layer 21. Following prior work~\cite{vojivr2024pixood, nayal2024likelihood}, we choose this foundation model as a backbone due to its feature representations, which encode rich contextual and semantic information. Since DINOv2 operates on image patches, it outputs feature maps $\mathbf{x} \in \mathbb{R}^{1024 \times H/14 \times W/14}$. 
\paragraph{Likelihood Ratio Estimator} For the estimator, we use a lightweight architecture consisting of three $1 \times 1$ convolutions and leaky ReLU activation, effectively acting as a fully connected \gls{nn} applied along the feature channel to classify individual feature vectors $\mathbf{x}_i \in \mathbb{R}^{1024}$. This design mitigates neighbouring \gls{id} pixels from influencing the classification of nearby \gls{ood} pixels, which is particularly important for detecting fine-grained or spatially small \gls{ood} regions. For the standard estimator, we use the same architecture, but with sigmoid output activation and trained with \gls{bce}. Both models are trained for 10 epochs with the Adam optimiser~\cite{kingma2014adam} and a learning rate of $2\cdot10^{-5}$.%This design prevents neighbouring \gls{id} pixels from influencing the classification of nearby \gls{ood} pixels, which is particularly important for detecting fine-grained or spatially small \gls{ood} regions. For the standard estimator, we use the same architecture, but with sigmoid output activation and trained with \gls{bce}. Both models are trained for 10 epochs with the Adam optimiser~\cite{kingma2014adam} and a learning rate of $2\cdot10^{-5}$.

\paragraph{\gls{ood} Score} We use the likelihood ratio from Eq.~\ref{eq:lr_estimator} as a scoring function, which for our proposed method evaluates to 
\begin{equation}
    \frac{p_{\theta}(y_i^{\text{out}}=1|\mathbf{x}_i)}{1 - p_{\theta}(y_i^{\text{out}}=1|\mathbf{x}_i)} = \frac{\alpha_{i1}}{S_i} \Big/ \frac{\alpha_{i0}}{S_i} = \frac{\alpha_{i1}}{\alpha_{i0}}.
\end{equation}
Following standard practice, we upscale the final \gls{ood} scores to match the original image resolution and apply Gaussian blurring with $\sigma=1$.

\subsubsection{Quantitative Results}
\label{sec:exp:quantitative}
\newcommand{\cmark}{\ding{51}}%
\newcommand{\xmark}{\ding{55}}%

\definecolor{firstcolor}{rgb}{0.0, 0.5, 0.0}
\definecolor{secondcolor}{rgb}{0.0, 0.125, 0.5}
%\definecolor{cellcolorhl}{rgb}{0.125, 1.0, 0.0}
\definecolor{cellcolorhl}{rgb}{0.2, 1.0, 0.20}
\newcommand{\first}{{\color{firstcolor!90}\ding{182}}{ }}%182,192,202
\newcommand{\second}{{\color{secondcolor!90}\ding{183}}{ }}%183,193,203
\newcommand{\phm}{\phantom{\first}}
\newcommand{\selectedcell}{\cellcolor{cellcolorhl!10}}

\begin{table*}[t]
%\caption{\textbf{Quantitative results.} Comparison with the state-of-the-art on the official SMIYC~\cite{chan2021segmentmeifyoucan} benchmark, Road Anomaly~\cite{lis2019detecting} and Fishyscapes Static~\cite{blum2021fishyscapes} validation set, along with the average performance across all five benchmark datasets. Methods with results reported on fewer than four datasets are reported in \textcolor{gray}{gray}. The best result for each dataset is highlighted in \textbf{bold}, and the second-best is \underline{underlined}.}
\caption{\textbf{Quantitative results.} Comparison with the state-of-the-art on the official SMIYC~\cite{chan2021segmentmeifyoucan} benchmark, Road Anomaly~\cite{lis2019detecting} and Fishyscapes Static~\cite{blum2021fishyscapes} validation set, along with the average performance across all five benchmark datasets. The best result for each dataset is highlighted in \textbf{bold}, and the second-best is \underline{underlined}. Notably, UEM~\cite{nayal2024likelihood} and PixOOD~\cite{vojivr2024pixood} also use a DINOv2 backbone.}
\label{tab:exp:quantitative}
\centering\resizebox{0.95\textwidth}{!}{
\begin{tabular}{lrrrrrrrrrr|rr}
\toprule

\multirow{2}{*}{Method} & \multicolumn{2}{c}{Anomaly Track} & \multicolumn{2}{c}{Obstacle Track} & \multicolumn{2}{c}{LaF NoKnown} & \multicolumn{2}{c}{Road Anomaly} & \multicolumn{2}{c}{FS Static} & \multicolumn{2}{c}{Average}\\

%\cmidrule(lr){3-5}
%\cmidrule(lr){6-8}
%\cmidrule(lr){9-11}
\cmidrule(lr){2-3}
\cmidrule(lr){4-5}
\cmidrule(lr){6-7}
\cmidrule(lr){8-9}
\cmidrule(lr){10-11}
\cmidrule(lr){12-13}
&
\multicolumn{1}{c}{AP $\uparrow$}& \multicolumn{1}{c}{FPR $\downarrow$}&
\multicolumn{1}{c}{AP $\uparrow$}& \multicolumn{1}{c}{FPR $\downarrow$}&
\multicolumn{1}{c}{AP $\uparrow$}& \multicolumn{1}{c}{FPR $\downarrow$}&
\multicolumn{1}{c}{AP $\uparrow$}& \multicolumn{1}{c}{FPR $\downarrow$}&
\multicolumn{1}{c}{AP $\uparrow$}& \multicolumn{1}{c}{FPR $\downarrow$}&
\multicolumn{1}{c}{$\overline{\text{AP}}$ $\uparrow$}& \multicolumn{1}{c}{$\overline{\text{FPR}}$ $\downarrow$}\\

\midrule

SynBoost~\cite{di2021pixel}      & 56.44         & 61.86 & 71.34 & 3.15 & 81.71 & 4.64 & -- & -- & 72.59 & 18.75 & 70.52 & 22.10\\
%ATTA~\cite{gao2023atta}                 & 67.04         & 31.57 & 76.46 & 2.81 & --    & -- \\
Maximized Entropy~\cite{chan2021entropy} & 85.47         & 15.00 & 85.07 & 0.75  & 77.90 & 9.70 & -- & -- & 81.00 & 5.00 & 82.36 & 7.61\\
PEBAL~\cite{tian2022pixel}        & 49.14         & 40.82 & 4.98  & 12.68 & --    & -- & 62.37 & 28.29 & 82.73 & 6.81 & 49.81 & 22.15\\
DenseHybrid~\cite{grcic2022densehybrid}          & 77.96         & 9.81  & 87.08 & 0.24  & 78.67 & 2.12 & -- & -- & -- & -- & 81.24 & 4.06 \\
DaCUP~\cite{vojivr2023image}            & --            & --    & 81.50 & 1.13  & 81.37 & 7.36 & -- & -- & -- & -- & 81.44 & 4.25 \\
RbA~\cite{nayal2023rba}                & 94.46  & 4.60  & \textbf{95.12} & \textbf{0.08}  & --    & -- & 85.42 & 6.92 & -- & -- & \underline{91.67} & 3.87  \\
EAM~\cite{grcic2023advantages}              &  93.75 & \underline{4.09}  & 92.87 & 0.52  & --    & -- & 69.40 & 7.70 & \underline{96.00} & \underline{0.30} & 88.51 & \underline{3.15} \\
NFlowJS~\cite{grcic2024dense}           & 56.92         & 34.71 & 85.55 & 50.36 & \textbf{89.28} & \textbf{0.65} & -- & -- & -- & -- & 77.25 & 28.57\\
Mask2Anomaly~\cite{rai2023unmasking}                       & 88.72  & 14.63 & 93.22    & 0.20  & --    & --  & 79.70 & 13.45 & 95.20 & 0.82 & 89.66 & 5.18\\
RPL\texttt{+}CoroCL~\cite{liu2023residual}         & 83.49         & 11.68 & 85.93 & 0.58 & --    & -- & 71.60 & 17.74 & 92.46 & 0.85 & 83.87 & 7.71 \\
Maskomaly~\cite{ackermann2023maskomaly}                & 93.35  & 6.87  & -- & -- & --   & -- & 70.90 & 11.90 & 69.50 & 14.40 & 77.92 & 11.06 \\
%ODIN~\cite{liang2018enhancing}          & 33.06         & 71.68 & 22.12 & 15.28 & 52.93 & 30.04 \\
%JSRNet~\cite{Vojir_2021_ICCV}           & 33.64         & 43.85 & 28.09 & 28.86 & 74.17 & 6.59  \\
%Image Resynthesis~\cite{Lis_2019_ICCV}  & 52.28         & 25.93 & 37.71 & 4.70  & 57.08 & 8.82  \\
cDNP~\cite{galesso2023far}           & 88.90  & 11.42 & --    & --  & --    & -- & 85.60 & 9.80 & -- & -- & 87.25 & 10.61 \\
Road Inpainting~\cite{lis2023detecting} & --            & --    & 54.14 & 47.12 & 82.93 & 35.75 & -- & -- & -- & -- & 68.54 & 41.44 \\
%ObsNet~\cite{besnier2021trigger}        & 75.44         & 26.69 & --    & --    & --    & --    \\
CSL~\cite{zhang2024csl}  & 80.08         & 7.16  & 87.10 & 0.67 & -- & -- & 61.38 & 43.80 & -- & -- & 76.19 & 17.88\\
PixOOD~\cite{vojivr2024pixood}                          & 68.88         & 54.33 &  88.90 & 0.30  & \underline{85.07} & 4.46 &  \textbf{96.39} &\textbf{ 4.30} & -- & -- & 84.31 & 15.85\\
UNO~\cite{delic2024outlier}                             & \textbf{96.33}  & \textbf{1.98} & 93.19    & 0.16  & --    & -- & 85.50 & 7.40 & \textbf{98.00} & \textbf{0.04} & \textbf{94.24} & 3.52\\
UEM~\cite{nayal2024likelihood}                             & \underline{95.60}  & 4.70 &  \underline{94.38}    & \underline{0.10}  & 81.04    & \underline{1.45} & 90.94    & 8.03 & -- & -- & 90.49 & 3.57 \\
%\pixood ({\bf Ours})                    & 68.88         & 54.33 & 19.82 &\first 88.90 &\first 0.30  & \second 50.82 &\second 85.07 &\second 4.46  & 44.41\\

\cmidrule(lr){1-13}

\rowcolor{gray!10}
BCE Baseline                   & 92.64 & 7.74 & 86.58 & 0.53 & 73.67 & 6.63 & 91.26 & 5.9 & 92.92 & 1.23 & 87.41 & 4.41\\

\rowcolor{gray!10}
EDL (ours)                    & 94.19 & 5.82 & 91.07 & 0.19 & 80.90 & 1.20 & \underline{92.95} & \underline{4.81} & 95.44 & 0.49 & 90.91 & \textbf{2.50}\\

\bottomrule
\end{tabular}
}
%\label{tab:smiyc_results_main}
\end{table*}%

Tab.~\ref{tab:exp:quantitative} shows the results on the official SMIYC benchmark, Road Anomaly and Fishyscapes Static datasets. Across all benchmarks, our proposed method (EDL) consistently outperforms the standard estimator (BCE). The most notable improvement in both AP and FPR is observed on the LaF NoKnown dataset, which contains real-world obstacles such as boxes and pallets that differ substantially from the semantically disjoint COCO objects used during training. This highlights the robustness of our approach and underscores the importance of effectively distinguishing intrinsic uncertainty from true \gls{ood} objects, especially when synthetic proxy features deviate from real-world \gls{ood} instances.

Recent state-of-the-art methods such as EAM~\cite{grcic2023advantages} and UNO~\cite{delic2024outlier} have extended their training setups by incorporating additional \gls{id} data from the Mapillary Vistas dataset~\cite{neuhold2017mapillary}. In addition, both methods leverage richer outlier supervision by using the ADE20K dataset~\cite{zhou2017scene}, which covers a broader and more diverse set of semantic classes compared to COCO. Since training data significantly affects performance, we restrict our setup to Cityscapes and COCO to ensure fair comparisons with prior work. However, despite these differences in the supervision, our approach achieves competitive results on individual benchmarks, especially, on Road Anomaly where we outperform the above methods and are the runner-up to PixOOD~\cite{vojivr2024pixood}.

Compared to prior approaches, our method achieves the lowest average false positive rate $\overline{\text{FPR}}$ across all five benchmarks. While other methods show strong performance on specific datasets, our method provides more balanced and consistently strong results across diverse open-set driving scenarios.

\subsubsection{Qualitative Results}
Fig.~\ref{fig:teaser} presents visual examples from our method. Due to its pixel-wise nature, our approach produces more fragmented \gls{ood} maps compared to recent mask-based methods~\cite{delic2024outlier, nayal2023rba}. Since our method detects unknown objects w.r.t. the given \gls{id} dataset, the model under-segments tyres and parts of trailers because they are present in \gls{id} objects from Cityscapes. Nevertheless, our method consistently identifies truly unknown objects, which is ultimately more critical than achieving visually complete but potentially incorrect segmentations. Post-processing techniques, such as those proposed in~\cite{zhao2024segment}, can be applied to further refine the output and improve segmentation precision.

\subsection{Ablation Study}
We investigate the impact of feature representations by conducting an ablation study on the choice of backbone layers. Furthermore, we compare the behaviour of our evidential model to that of a standard classifier, evaluating whether the observations from Sec.~\ref{sec:exp:univariate_gaussians} generalise to pixel-wise \gls{ood} detection.
\begin{figure}%[htbp]
  \centering
  \includegraphics[width=\columnwidth]{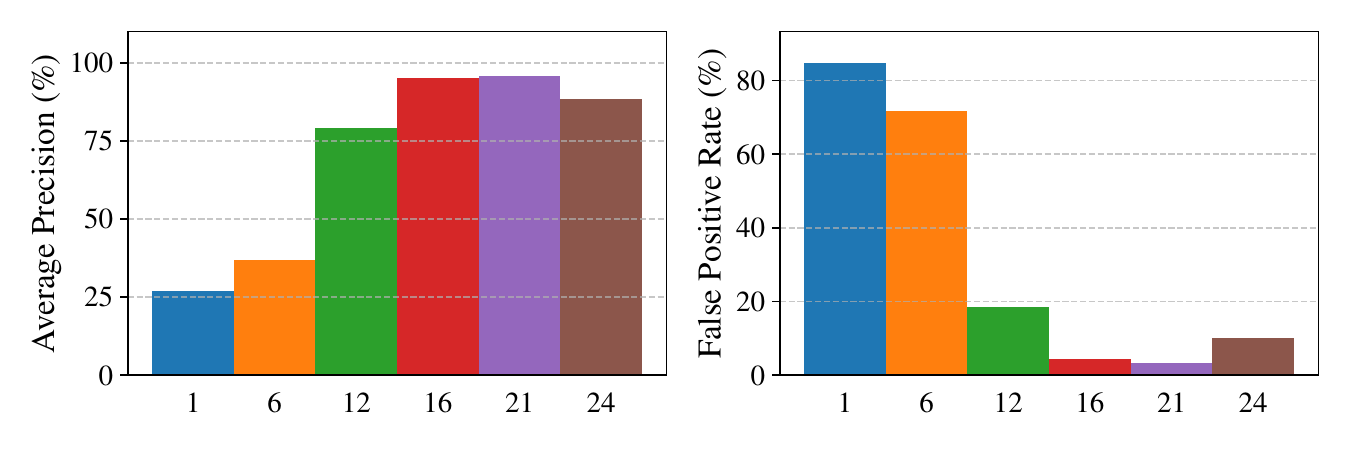}
   \vspace{-2.5em}
  \caption{\textbf{DINOv2 layer ablation study.} Comparison of models trained using features from different layers $l \in \{1, 6, 12, 16, 21, 24\}$ of DINOv2. Performance is reported in terms of average precision ($\uparrow$) and false positive rate ($\downarrow$) on the SMIYC Anomaly Track validation set.}
  \label{fig:exp:ablation_study}
\end{figure}
\subsubsection{Feature Impact} In Fig.~\ref{fig:exp:ablation_study}, we evaluate the performance of our method using features extracted from a representative subset of the 24 layers of DINOv2, spanning early, middle, and late stages of the network. Our results indicate that features from earlier layers ($l \in \{1,6,12\}$) lack sufficient semantic abstraction, while those from the final layer ($l=24$) are less informative for reliable pixel-wise \gls{ood} detection. In contrast, mid- to high-level features ($l \in \{16,21\}$) provide the best performance, effectively balancing detailed spatial with high-level semantic information.
%In Fig.~\ref{fig:exp:ablation_study}, we compare the performance of our method when trained on features extracted from a representative subset of the 24 layers of DINOv2, covering early, middle, and late stages. Our results show that features from layer 21 yield the best performance, striking a balance between mid-level details and high-level semantic abstraction.
\iffalse\textcolor{blue}{
\paragraph{Backbone Influence} Perform experiments with a \gls{cnn} DeepLabV3+. Following~\cite{liu2023residual}
\input{tables/smiyc_results_ablation}Unlike methods relying on masked-based, our apporach is applicable to CNN-based backbones.
}\fi
\subsubsection{Extrapolation Analysis} 
%We compare the behaviour of our evidential model to that of a standard classifier on the Fishyscapes Static dataset, evaluating whether the observations from Sec.~\ref{sec:exp:univariate_gaussians} generalize to this scenario.
\begin{figure}[t]%[htbp]
  \centering
  \includegraphics[width=\columnwidth]{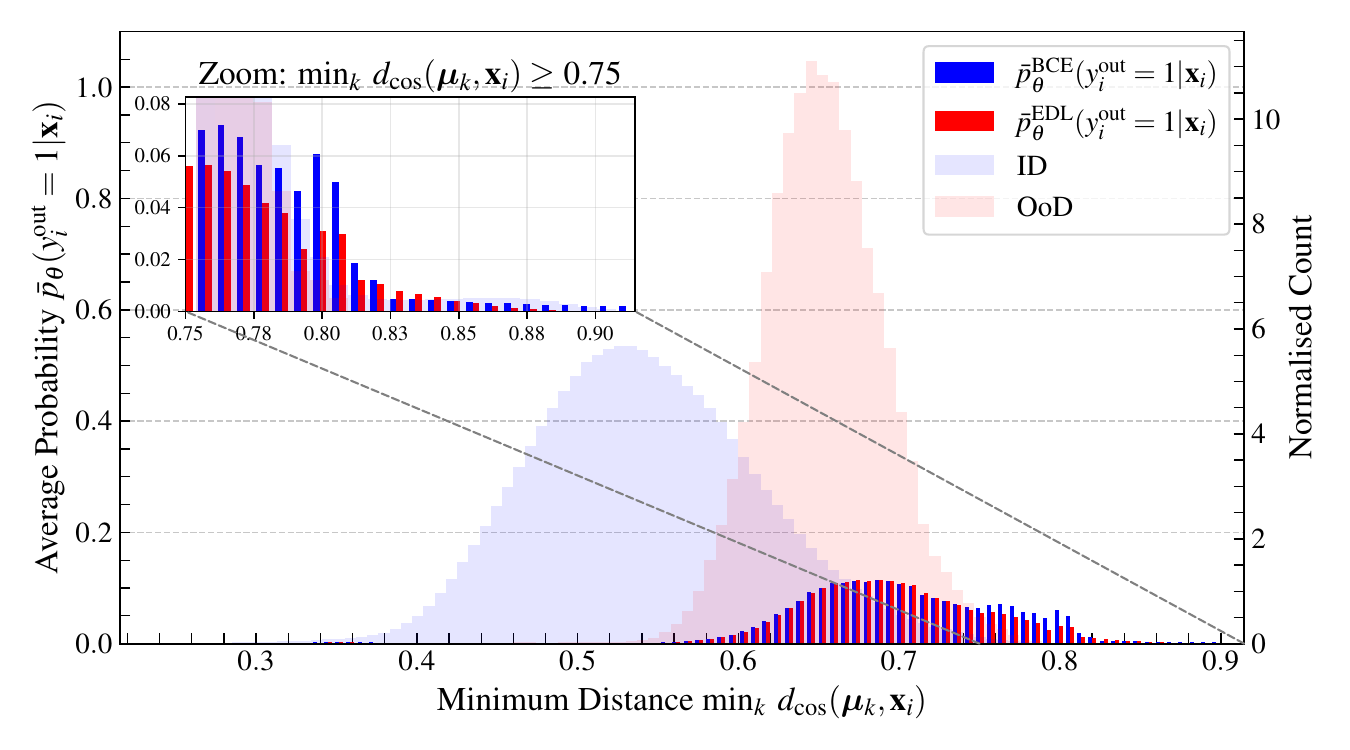}
   \vspace{-2.5em}
  \caption{\textbf{Visualisation of overconfident extrapolation.} Comparison between an evidential classifier (EDL) and a standard binary classifier (BCE) on the Fishyscapes Static dataset. The plot shows the average probability $\bar{p}_{\theta}(y^{\text{out}}=1|\mathbf{x}_i)$ over the cosine distance $d_{\mathrm{cos}}(\boldsymbol{\mu}_k, \mathbf{x}_i)$ between a feature vector $\mathbf{x}_i$ and the nearest class mean feature vector $\boldsymbol{\mu}_k$ from the training set. For reference, shaded blue and red areas indicate the density of distances for \gls{id} and \gls{ood} features, respectively.}
  \label{fig:exp:overconfidence}
\end{figure}

In order to compare the behaviour of our evidential
model to that of a standard classifier, we extract feature maps from each image in the Fishyscapes Static dataset using the DINOv2 backbone and obtain corresponding feature vectors $\mathbf{x}_i$. Due to the high dimensionality of these vectors, direct estimation of the feature space density is unreliable. Moreover, DINOv2 features encode semantic information in their orientation or angular relationships \cite{oquab2023dinov2}. Therefore, instead of estimating density, we quantify feature similarity via the cosine distance
\begin{equation}
    d_{\mathrm{cos}}(\mathbf{x}_i, \boldsymbol{\mu}_k) = 1 - \frac{\mathbf{x}_i \cdot \boldsymbol{\mu}_k}{\left\|\mathbf{x}_i \right\| \left\|\boldsymbol{\mu}_k \right\|},
\end{equation}
where $\boldsymbol{\mu}_k$ denotes the mean feature vector of class $k$ in the training set. For each feature vector $\mathbf{x}_i$, we first normalise it to unit length and compute its cosine distance to the nearest class mean $\boldsymbol{\mu}_k$. This distance serves as a proxy for feature novelty, where larger distances indicate less similarity to the training distribution and potentially highlight regions where extrapolation occurs. 

In Fig.~\ref{fig:exp:overconfidence}, we group feature vectors by their minimum cosine distance (binned) and plot the average predicted probability $\bar{p}_{\theta}(y_i^{\text{out}}=1|\mathbf{x}_i)$ for each bin. Consistent with our observations in Sec.~\ref{sec:exp:univariate_gaussians}, the standard classifier exhibits \textit{overconfident behaviour}, assigning higher predicted probabilities even to test features that are highly dissimilar to the training data. In contrast, our proposed evidential model appropriately reduces predictive probability in these regions, effectively mitigating overconfident extrapolation. This behaviour is also quantitatively reflected in the overall lower FPR of our estimator compared to the standard estimator in Tab. \ref{tab:exp:quantitative}.

%------------------------------------------------------------------------
\section{Conclusion}
We presented an uncertainty-aware likelihood ratio estimator for pixel-wise out-of-distribution detection. Our method addresses limitations of existing approaches by leveraging evidential deep learning to explicitly model uncertainty arising from long-tailed class distributions and imperfect synthetic outliers. By predicting probability distributions instead of point estimates, our approach improves robustness to both rare in-distribution classes and distributional shifts in proxy \gls{ood} features. Through comprehensive evaluation across five standard benchmarks, including the official Segment-Me-If-You-Can benchmark, our method achieves the lowest average false positive rate (2.5\%) among state-of-the-art while maintaining high average precision (90.91\%). Importantly, this performance is achieved with negligible computational overhead and without degrading \gls{id} segmentation accuracy. These results demonstrate that accounting for intrinsic uncertainty enables more effective use of outlier exposure and reduces false detections in complex real-world driving scenarios. Future work may extend our approach toward a general framework for statistical hypothesis testing with likelihood ratios under distributional uncertainty.

{
    \small
    \bibliographystyle{ieeenat_fullname}
    \bibliography{references}
}

\end{document}